\pdfoutput=1

\documentclass{article}

\usepackage{microtype}
\usepackage{graphicx}
\usepackage{subcaption}
\usepackage{booktabs} 
\usepackage{makecell}
\usepackage{hyperref}

\usepackage{algpseudocode}

\usepackage{tabularx}
\usepackage{caption}
\usepackage{xcolor}
\usepackage{listings}

\lstdefinelanguage{json}{
    basicstyle=\normalfont\ttfamily,
    numbers=left,
    numberstyle=\scriptsize,
    breaklines=true,
    frame=lines,
    backgroundcolor=\color{gray!10},
    showstringspaces=false,
    string=[db]{"},
    stringstyle=\color{green!50!black},
    morestring=[s][\color{black}]{\ \ "}{":},
    keywordstyle=\color{blue},
    keywords={true,false,null},
    literate=
     *{0}{{{\color{red}0}}}{1}
      {1}{{{\color{red}1}}}{1}
      {2}{{{\color{red}2}}}{1}
      {3}{{{\color{red}3}}}{1}
      {4}{{{\color{red}4}}}{1}
      {5}{{{\color{red}5}}}{1}
      {6}{{{\color{red}6}}}{1}
      {7}{{{\color{red}7}}}{1}
      {8}{{{\color{red}8}}}{1}
      {9}{{{\color{red}9}}}{1}
      {.}{{{\color{red}.}}}{1}
      {:}{{{\color{gray}{:}}}}{1}
      {,}{{{\color{gray}{,}}}}{1}
      {\{}{{{\color{gray}{\{}}}}{1}
      {\}}{{{\color{gray}{\}}}}}{1}
      {[}{{{\color{gray}{[}}}}{1}
      {]}{{{\color{gray}{]}}}}{1},
}

\usepackage[preprint]{icml2026}

\usepackage{amsmath}
\usepackage{amssymb}
\usepackage{mathtools}
\usepackage{amsthm}
\usepackage[T1]{fontenc}

\lstdefinestyle{prompt}{
  basicstyle=\ttfamily\small,
  columns=fullflexible,
  breaklines=true,
  frame=single,
  framerule=0.4pt,
  rulecolor=\color{black!20},
  backgroundcolor=\color{black!2},
  xleftmargin=0.6em,
  xrightmargin=0.6em,
  aboveskip=0.6em,
  belowskip=0.6em
}

\usepackage[capitalize,noabbrev]{cleveref}

\theoremstyle{plain}

\theoremstyle{definition}

\theoremstyle{remark}

\usepackage[disable,textsize=tiny]{todonotes}

\icmltitlerunning{Self-Reasoning Agentic Framework for Narrative Product Grid Collage Generation}

\begin{document}

\twocolumn[{
  \icmltitle{Self-Reasoning Agentic Framework \\ 
  for Narrative Product Grid-Collage Generation}



  \icmlsetsymbol{equal}{*}

  \begin{icmlauthorlist}
    \icmlauthor{Minyan Luo}{CASIA}
    \icmlauthor{Yuxin Zhang}{UCAS}
    \icmlauthor{Yifei Li}{THU}
    \icmlauthor{Xincan Wang}{STA}
    \icmlauthor{Fuzhang Wu}{CASIS}
    \icmlauthor{Tong-Yee Lee}{CKU}
    \icmlauthor{Oliver Deussen}{Konstanz}
    \icmlauthor{Weiming Dong}{CASIA}\!\textsuperscript{*}

  \end{icmlauthorlist}

  \icmlaffiliation{UCAS}{ByteDance}

  \icmlaffiliation{CASIA}{MAIS,Institute of Automation, Chinese Academy of Sciences}
  \icmlaffiliation{THU}{Tsinghua University}
  \icmlaffiliation{STA}{Shanghai Theatre Academy}
  \icmlaffiliation{CASIS}{Institute of Software, Chinese Academy of Sciences}
  \icmlaffiliation{CKU}{National Cheng Kung University}
  \icmlaffiliation{Konstanz}{University of Konstanz}

  \icmlcorrespondingauthor{Weiming Dong}{weiming.dong@ia.ac.cn}

  \icmlkeywords{Machine Learning, ICML}
\begin{center}
    \centering
    \captionsetup{type=figure}
    \includegraphics[width=\linewidth]{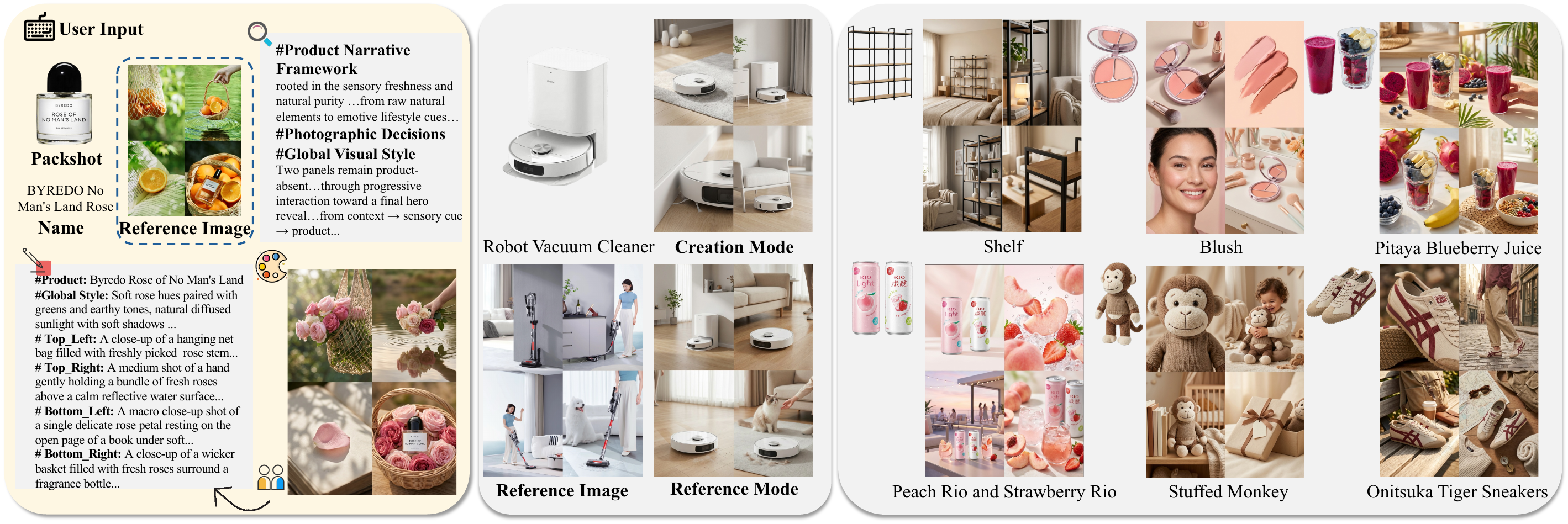}


    \caption{Examples of narrative product grid collage generation. Rather than merely depicting a product in isolation, we generate grid collages that incorporate narrative elements, establishing associations with functionality, affordances, or lifestyle enhancement. Our method can further transfer the photographic direction from a reference image while being specifically tailored to the target product, thereby generating a coherent and product-centric visual narrative.}
    \label{fig_teaser}
    \vspace{-2mm}
\end{center}

  \vskip 0.2in
  }
]



\printAffiliationsAndNotice{}  


\begin{abstract}
Narrative-driven product photography has become a prevalent paradigm in modern marketing, as coherent visual storytelling helps convey product value and establishes emotional engagement with consumers.
However, existing image generation methods do not support structured narrative planning or cross-panel coordination, often resulting in weak storytelling and visual incoherence.
In practice, narrative product photography is commonly presented as multi-grid collages, where multiple views or scenes jointly communicate a product narrative. To ensure visual consistency across grids and aesthetic harmony of the overall composition, we generate the collage as a single unified image rather than composing independently synthesized panels. We propose a self-reasoning agentic framework for narrative product grid collage generation. Given a product packshot and its name, the system first constructs a \textit{Product Narrative Framework} that explicitly represents the product’s identity, usage context, and situational environment, and translates it into complementary grids governed by a shared visual style. Constraint-aware prompts are then compiled and fed to a generation model that synthesizes the collage jointly. The generated output is evaluated on both content validity and photography quality, with explicit gates determining whether to proceed or refine. When evaluation fails, the system performs failure attribution and applies targeted refinement, enabling progressive improvement through iterative self-reflection. Experiments demonstrate that our framework consistently improves aesthetic quality, narrative richness, and visual coherence, compared to direct prompting baselines.

\end{abstract}
\section{Introduction}

\begin{figure*}
\centering
\includegraphics[width=1.0\linewidth]{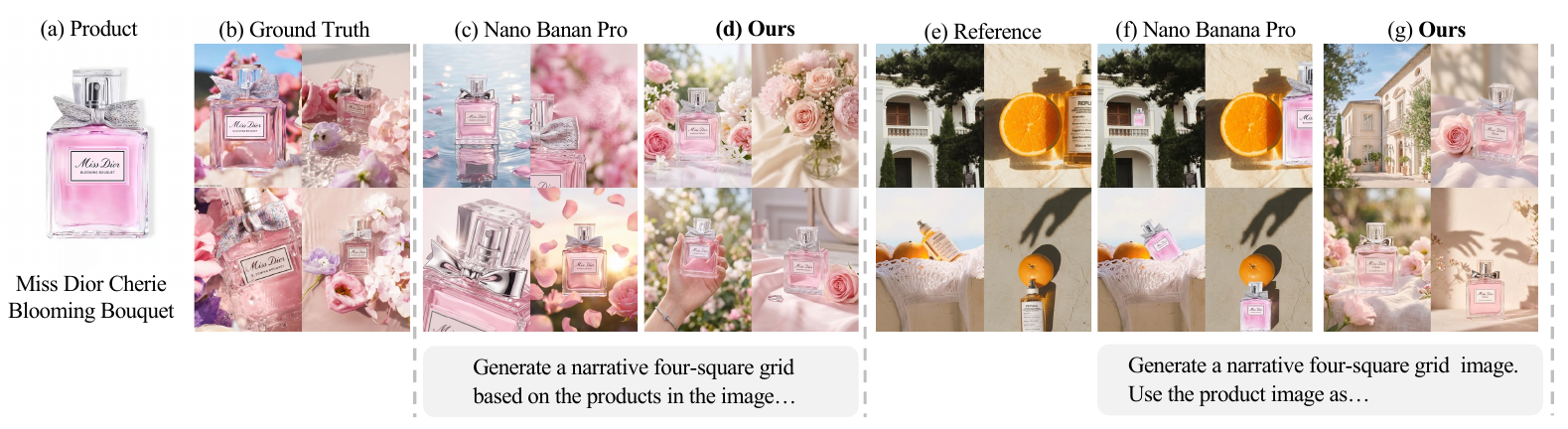}
\caption{With identical inputs, our method yields improved aesthetic quality, stronger narrative richness, higher visual coherence than Nano Banana Pro direct generation.
(a) Product packshot. (b) Ground-truth four-panel campaign grid staged and photographed by professional photographers.
\textbf{Creation mode:} (c) Nano Banana Pro direct output using the in-figure prompt caption; (d) our result built on Nano Banana Pro under the same prompt.
\textbf{Reference mode:} (e) reference image; (f) Nano Banana Pro direct output using the in-figure prompt caption; (g) our result built on Nano Banana Pro.
}
\vspace{-4mm}
\label{fig_compare}
\end{figure*}

If a photo captures a moment, product photography tells a story. 
Product photography is essential for persuasive branding~\cite{azimi2012userresponse,cheng2012multimedia}. In fast-paced e-commerce and social media environments (e.g., X, Instagram, and RedNote), users often decide to engage with a post within milliseconds. Consequently, a cover image must immediately capture attention and convey a structured narrative.
To meet this demand, product collages (e.g., $2 \times 2$ or $3 \times 3$ layouts) have emerged as a dominant representational form. By juxtaposing multiple viewpoints, scenarios, or narrative fragments, these layouts deliver high-density information and spatialized narratives while reducing cognitive load. 
Importantly, such layouts are not merely aggregations of multiple images but constitute a visual organization strategy with explicit narrative logic.
As shown in Figure~\ref{fig_teaser}, integrating narrative frameworks into collages unifies product attributes, usage contexts, environmental settings, and affective cues into a coherent visual structure.

Despite 62\% of marketers now employing generative AI for image creation~\cite{salesforce2025ai}, current tools~\cite{deepmind_gemini3pro,seedream2025seedream,labs2025flux,recraft2025ad} remain ill-suited for narrative product imagery. Achieving professional results requires complex, manually crafted prompts; otherwise, outputs lack narrative depth. This limitation is particularly evident in multi-grid scenarios, where visual homogeneity undermines storytelling (see Figure~\ref{fig_compare}). Existing e-commerce image synthesis methods~\cite{chen2025ctr,feng2025forage,yang2024creative,du2024towards} focus on single-image optimization such as background replacement, preserving product details but failing to generate diverse viewpoints for effective storytelling. Story visualization~\cite{zhou2025agentstory,yang2025seedstory,zheng2025contextualstory,he2025dreamstory} and animated storytelling methods~\cite{li2024animdirector,zhang2025anime,shi2025animaker} address sequential structures where frames evolve over time. However, no prior work has addressed the challenge of product collage generation: parallel creation of visually diverse yet thematically unified grid sets that collectively convey brand narratives while ensuring aesthetic quality and visual coherence.

To address these limitations, we propose a \textit{Self-Reasoning Agentic Framework} for narrative product grid collage generation. Our primary insight is that effective visual storytelling requires separating \textit{narrative reasoning} from \textit{pixel synthesis}. By constructing a \textit{Product Narrative Framework}, we introduce a structured intermediate representation capturing the product's ontology, usage context, and sensory atmosphere before generation. This decoupling enables reasoning about ``implicit narratives'', i.e., scenes where the product is thematically pervasive even if physically absent, enriching storytelling beyond literal object depiction.
We address structural generalization through two operating modes. In \textit{Creation Mode}, the framework explores creative associations, constructing coherent multi-grid narratives with complementary scenes and stylistic variations. In \textit{Reference Mode}, we extract high-level photographic decisions (e.g., shot scale, composition, interaction) from exemplar advertisements, adapting them to target products without copying pixel-level appearance. To enforce visual coherence, we employ joint synthesis that generates the multi-grid collage as a unified visual field, avoiding the ``visual drift'' of sequential generation methods.
Finally, we integrate a closed-loop \textit{Ideation-Generation-Critique} mechanism, motivated by the insight that MLLMs exhibit stronger capabilities in critical evaluation than one-shot generation. An autonomous critic evaluates outputs against the \textit{Product Narrative Framework} for content validity (what to shoot) and execution validity (how to shoot), triggering corresponding revisions when criteria fall short.

Our contributions are summarized as follows:
\begin{itemize}
    \item We introduce an agentic framework for narrative product grid collage generation that bridges the gap between pixel-level synthesis and structured design reasoning, supporting both reference-free creation and reference-guided transfer.
\item We propose the concept of a \textit{Product Narrative Framework} as an explicit narrative foundation, and design an iterative \textit{Ideation–Generation–Critique} loop with hierarchical gating for autonomous reflection.
\item We conduct comprehensive evaluations, demonstrating substantial advancement of our framework in generating grid collage product images with rich brand narratives, high aesthetic quality, and visual coherence.
\end{itemize}
\section{Related Work}

\paragraph{Text-to-Image Generation.}
Diffusion models have become the dominant paradigm for text-conditioned image synthesis. Models such as DALL$\cdot$E~\cite{ramesh2022hierarchical}, Imagen~\cite{saharia2022photorealistic}, and Stable Diffusion~\cite{rombach2022high} achieve remarkable fidelity. Recent works address complex constraints: Qwen-Image~\cite{wu2025qwen} demonstrates breakthroughs in text rendering and editing, FLUX.1 Kontext~\cite{labs2025flux} introduces in-context generation for iterative refinement, and Gemini-3-Pro Image~\cite{deepmind_gemini3pro} shows strong instruction-following and world knowledge integration. Despite these advances, most models operate in a single-image paradigm without mechanisms for narrative reasoning or cross-image consistency.

\begin{figure*}
\centering
\includegraphics[width=1\linewidth]{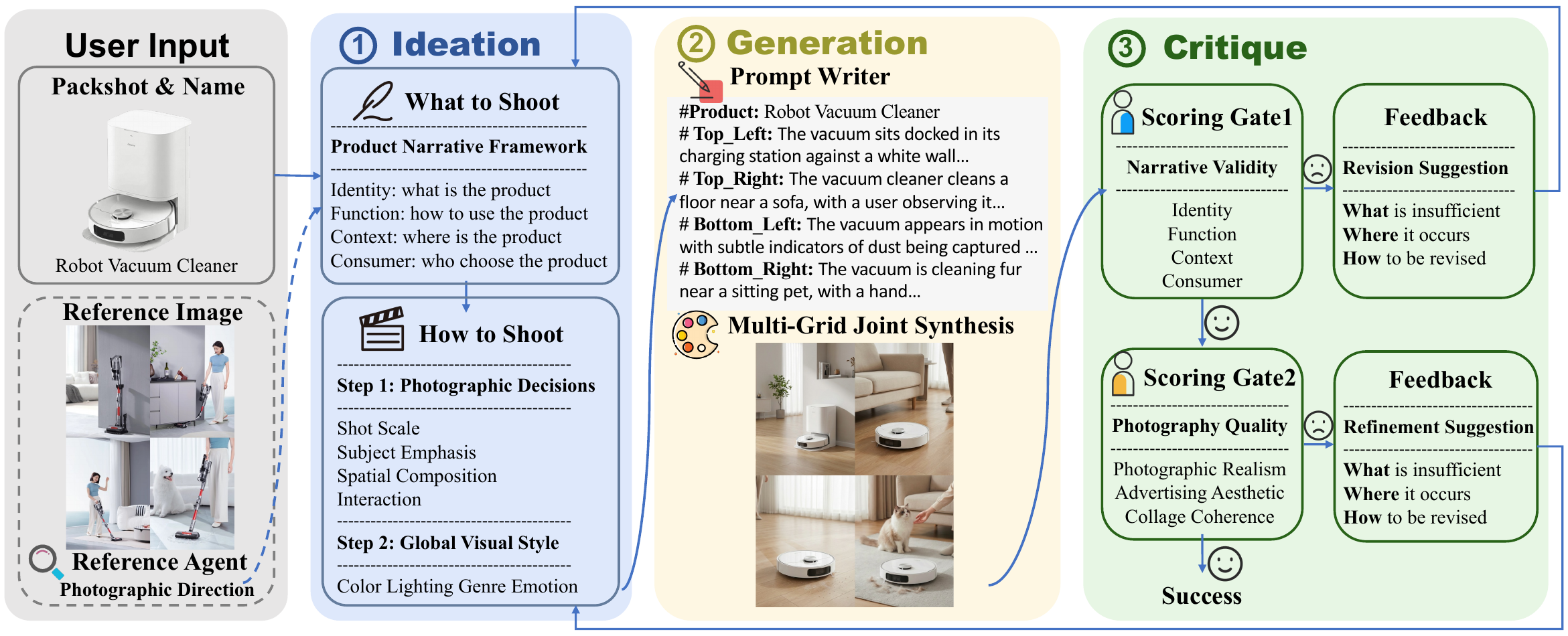}
\caption{The architecture of our proposed self-reasoning agentic framework for narrative product grid collage generation.
}
\vspace{-4mm}
\label{fig_pipeline}
\end{figure*}

\paragraph{Agent-based Visual Generation}
To move beyond one-shot generation, researchers have integrated LLM-based controllers with image synthesis models. Visual ChatGPT~\cite{wu2023visual} orchestrates multiple vision models via dialogue agents for multi-step tasks. Self-correction pipelines such as SLD~\cite{wu2024self} iteratively refine diffusion outputs by adjusting prompt mismatches. Idea2Img~\cite{yang2024idea2img} employs GPT-4V in iterative ideation–generation loops, while GenArtist~\cite{wang2024genartist} decomposes complex requests into sub-problems with specialist models. For visual storytelling, StoryAgent~\cite{hu2024storyagent} and MM-StoryAgent~\cite{xu2025mm} organize separate agents for scripting, generation, and synthesis. They demonstrate that multi-stage reasoning improves complex visual generation, yet most target single images or rely on predefined subtask sequences. We address this gap with a self-reasoning agent that plans and critiques visual narratives through an Ideation-Generation-Critique loop.


\paragraph{AIGC in E-commerce}
Generative AI is increasingly deployed in e-commerce to automate marketing creatives. A prominent thread studies \emph{product-background generation} via diffusion-based inpainting, where preserving product fidelity while producing aesthetic scenes is key. Representative methods leverage category/brand priors~\cite{wang2025generate}, enable prompt-free synthesis~\cite{cao2024product2img}, and incorporate visual references for multimodal control~\cite{zhao2025dreampainter}. DEPO~\cite{sun2025depo} adapts preference optimization to match human aesthetic judgments. Beyond backgrounds, \emph{layout-aware ad generation} methods such as PosterMaker~\cite{gao2025postermaker}, T-Stars-Poster~\cite{chen2025t}, and PosterVerse~\cite{liu2026posterverse} model end-to-end poster synthesis with explicit layout components. Complementary work extends to \emph{marketing language generation}~\cite{liu2025llms,quan2025crmagent}. Despite these advances, most systems optimize a \emph{single} asset and lack mechanisms for campaign-level narrative planning and cross-image coherence. Our method addresses this gap with a self-reasoning agent that iteratively plans and critiques narrative grid-collages for holistic story logic and visual consistency.

\section{Method}
\label{sec:method}

We propose a self-reasoning agentic framework designed to generate coherent product narratives in the form of grid collage. Formally, given a product packshot $I_p$, a product name $I_n$, and an optional reference image $I_r$, our system generates a multi-grid collage $\mathcal{C}$. 
As shown in Figure~\ref{fig_pipeline}, the framework orchestrates four specialized LLM-based agents: the \textbf{Reference Agent}($\mathcal{A}_{\text{ref}}$), the \textbf{Ideation Agent}($\mathcal{A}_{\text{idea}}$), the \textbf{Generation Agent} ($\mathcal{A}_{\text{gen}}$) and the \textbf{Critique Agent} ($\mathcal{A}_{\text{crit}}$). Algorithm~\ref{alg:pipeline_short} summarizes the overall pipeline.

\subsection{Ideation}
\label{sec:ideation}
Ideation phase is handled by $\mathcal{A}_{\text{idea}}$ and proceeds in two sequential steps: \textit{What to Shoot} and \textit{How to Shoot}. Implementation details are deferred to Appendix~\ref{app:creation_agent} and Appendix~\ref{app:reference_agent}.

\paragraph{What to Shoot}
This step focuses on what to shoot in a multi-grid collage by constructing a complete \textit{Product Narrative Framework($\mathcal{F}$)}. 
Rather than enumerating surface-level attributes, $\mathcal{F}$ captures the product’s underlying semantics, forming a coherent narrative logic that can motivate purchase~\cite{escalas2004narrative}.
Concretely, $\mathcal{F}$ is defined along four fundamental semantic dimensions:
\textbf{Identity ($\mathbf{f}_{\text{what}}$)} characterizes what the product fundamentally is, including its category, material composition, sensory attributes (e.g., color, texture, scent), and distinguishing features.
\textbf{Function} ($\mathbf{f}_{\text{how}}$) specifies what the product does and how it is used, describing typical usage patterns, interaction modes, and the actions or benefits the product enables.
\textbf{Context} ($\mathbf{f}_{\text{where}}$) situates the product within realistic environments and situations, capturing where it naturally appears and acquires meaning in daily life.
\textbf{Consumer ($\mathbf{f}_{\text{who}}$)} defines who the product is for and why they would choose it, emphasizing decision logic, needs, motivations, and lifestyle orientation.

Crucially, $\mathcal{F}$ supports \textit{implicit narratives}, in which the product may be physically absent yet thematically pervasive. Such scenes depicting raw ingredients, usage traces, or evocative environmental cues, enrich the narrative framework by stimulating viewer imagination and fostering deeper engagement~\cite{woodside2008consumers}.

\paragraph{How to Shoot}
This step translates the \textit{Product Narrative Framework} into \textit{Photographic Decisions}($\mathcal{P}$) and determines a \textit{Global Visual Style}($\mathcal{S}$) that jointly govern all grids.

\textit{Photographic Decisions} $\mathcal{P}$ define the intended photographic role of each grid, where
$\mathcal{P} = \{p_i\}_{i=1}^N$ and each $p_i$ corresponds to one grid in the collage. Specifically, $\mathcal{P}$ comprises four dimensions.
\textbf{Shot Scale ($\mathbf{s}_{\text{scale}}$)} characterizes the intended camera distance (e.g., macro, close-up, medium, wide), regulating perceptual intimacy and the level of visual detail.
\textbf{Subject Emphasis ($\mathbf{s}_{\text{subj}}$)} defines the hierarchy between primary and secondary elements, specifying how viewer attention is directed through focus, depth of field, and saliency control.
\textbf{Spatial Composition ($\mathbf{s}_{\text{comp}}$)} specifies the high-level geometric arrangement of elements, governing their relative placement,
scale, and visual balance. \textbf{Interaction Logic} ($\mathbf{s}_{\text{int}}$) defines how the product engages with its environment, or how the environment implies the product’s presence when it is not explicitly shown.

\textit{Global visual style} $\mathcal{S}$ acts as a campaign-level anchor, which constrains all grids to a shared visual world.
$\mathcal{S}$ defines global aesthetic attributes that are applied consistently across grids, spanning four dimensions: 
\textbf{Color} (palette selection, saturation level, and overall harmony), 
\textbf{Lighting} (illumination direction, contrast, and shadow–highlight behavior),
\textbf{Genre} (realism level, photographic treatment, and commercial polish),
and \textbf{Emotion} (affective atmosphere and mood consistency).

In \textbf{Reference Mode}, $\mathcal{A}_{\text{ref}}$ analyzes a reference image $I_r$ to extract transferable directions $\mathcal{D}_r$, which provide a high-level direction, implicitly guiding narrative framework, photographic decisions and global visual styles. $\mathcal{A}_{\text{idea}}$ is required to learn from this structural prior and generate a collage that follows similar narrative logic and composition, while remaining faithful to its own product identity. Details are deferred to Appendix~\ref{app:reference_agent}

\subsection{Generation}
\label{sec:Generation}

Generation phase transforms the \textit{Ideation} output $\mathcal{P} = \{ p_i \}_{i=1}^N$ and $\mathcal{S}$ into concrete visual outputs through prompt compilation and multi-grid joint synthesis.

\paragraph{Prompt Writer.}
The Generation Agent $\mathcal{A}_{\text{gen}}$ compiles a set of coordinated natural language prompts $\mathcal{PR} = \{pr_i\}_{i=1}^N$. Each prompt $pr_i$ integrates grid-specific decision derived from $p_i$ and style constraints from $\mathcal{S}$, ensuring cross-grid coherence at the textual level.
\paragraph{Multi-Grid Joint Synthesis.}
The Generation Agent $\mathcal{A}_{\text{gen}}$ produces the unified collage $\mathcal{C}$ conditioned on $\mathcal{PR}$, $I_p$, and $I_n$. Crucially, we employ a joint synthesis strategy rather than independent generation. This enforces global consistency across four grids.

\subsection{Critique}
\label{sec:Critique}


The critique phase functions as a closed-loop feedback mechanism implemented by $\mathcal{A}_{\text{crit}}$.
It employs a hierarchical, two-gate evaluation strategy that assigns quantitative scores to assess
\textit{narrative validity} and \textit{photographic quality}. When predefined thresholds are not met,
produces structured diagnostic feedback to guide targeted
revision or refinement of the generation process.

\paragraph{Gate 1: Narrative Validity}


We first assess whether generated grid-collage $\mathcal{C}$ successfully instantiates the intended \textit{product narrative framework}. $\mathcal{A}_{\text{crit}}$ assigns a narrative score $Score_{\text{narr}}(\mathcal{C}) \in [0,5]$.
Concretely, $E_{\text{narr}}$ evaluates $\mathcal{C}$ along four semantic dimensions:
\textbf{Identity} (what the product is): whether the grid clearly communicates the product category and defining attributes;
\textbf{Usage} (how it is used): whether the grid conveys the product's function and usage scenario;
\textbf{Context} (where/when it appears): whether the depicted scenes reflect natural settings and occasions for the product;
\textbf{Consumer} (for whom/why): whether the grid suggests the target audience and motivations for choosing the product.



If the aggregated score $Score_{\text{narr}}$ falls below a threshold $\tau_{\text{narr}}$, $\mathcal{A}_{\text{crit}}$ triggers \textit{Revision}, which operates at the narrative level to restructure and re-plan the product narrative framework.
Concretely, $\mathcal{A}_{\text{crit}}$ generates a structured diagnostic revision signal $Sug_{\text{narr}}$, formalized as a tuple $(\textbf{What}, \textbf{Where}, \textbf{How})$. 
Here, \textbf{what} specifies which aspect of the narrative framework breaks, \textbf{where} identifies the precise locus of the issue, and \textbf{how}  prescribes the corresponding revision directive.
Based on this feedback, the ideation agent selectively reconstructs the corresponding components of the narrative framework while freezing validated elements, before re-triggering generation.

\paragraph{Gate 2: Photography Quality}
While Gate~1 validates \textit{what} is depicted, Gate~2 evaluates
\textit{how well} the collage is rendered as a photographic advertisement.
$\mathcal{A}_{\text{crit}}$ by aggregating three perceptual criteria:
$Score_{\text{photo}}$ aggregates three perceptual metrics:
\textbf{Photographic Realism} evaluates physical plausibility and material credibility, such as correct geometry, realistic textures and reflections, and the absence of obvious generative artifacts (e.g., warped text, distorted objects).
\textbf{Advertising Aesthetic} assesses overall visual polish and commercial appeal, including composition quality, focal hierarchy, color harmony, and visual balance across grids.
\textbf{Collage Coherence} ensures the four grids feel like a unified campaign shot within a single session (aligned lighting direction, harmonized color temperature).

If the aggregated score $Score_{\text{photo}}$ falls below a threshold $\tau_{\text{photo}}$, $\mathcal{A}_{\text{crit}}$ triggers \textit{Refinement}, which operates at the photographic execution level to perform fine-grained adjustments while preserving the validated product narrative framework.
Concretely, $\mathcal{A}_{\text{crit}}$ outputs a structured refinement suggestion $Sug_{\text{photo}}$, which specifies
\textbf{what} aspect of photographic execution is insufficient,
\textbf{where} the issue manifests within the visual realization,
and \textbf{how} the photographic decisions or global visual style should be refined.
Overall, this hierarchical critique mechanism enables a closed-loop, self-reflective process in which semantic correctness and visual quality are disentangled, preventing the system from discarding coherent narrative structures due to pixel-level artifacts, while visual rendering is progressively refined.

\begin{algorithm}[t]
\caption{Self-Reflective Gated Pipeline for Product Collage Generation}
\label{alg:pipeline_short}
\begin{algorithmic}[1]
\Require Packshot $I_p$, product name $I_n$, optional reference $I_r$, thresholds $\tau_{\text{narr}},\tau_{\text{photo}}$, max iterations $K$
\Ensure Optimized grid collage $\mathcal{C}^*$

\State $\mathcal{D}_r \gets \emptyset$
\State $\mathcal{D}_r \gets \mathcal{A}_{\text{ref}}(I_r)$ if $I_r \neq \emptyset$

\State $(\mathcal{F},) \gets \mathcal{A}_{\text{idea}}(I_p, I_n, \mathcal{D}_r)$
\Comment{What to shoot}
\State $(\mathcal{P},\mathcal{S}) \gets \mathcal{A}_{\text{idea}}(I_p, I_n, \mathcal{F}, \mathcal{D}_r)$
\Comment{How to shoot}

\State $\mathcal{PR} \gets \mathcal{A}_{\text{gen}}(I_p,\mathcal{P}, \mathcal{S})$
\Comment{Prompt writer}
\State $\mathcal{C} \gets \mathcal{A}_{\text{gen}}(\mathcal{PR}, I_p)$
\Comment{Joint multi-grid generation}

\For{$k \gets 1$ to $K$}
    \State $Score_{\text{narr}} \gets \mathcal{A}_{\text{crit}}(I_p, I_n,\mathcal{C})$
    \State $Score_{\text{photo}} \gets \mathcal{A}_{\text{crit}}(I_p, I_n,\mathcal{C})$

    \If{$Score_{\text{narr}} < \tau_{\text{narr}}$}
        \State $Sug_{\text{narr}} \gets \mathcal{A}_{\text{crit}}(\mathcal{C}, Score_{\text{narr}})$
        \State $\mathcal{F} \gets \mathcal{A}_{\text{idea}}(I_p, I_n,\mathcal{F}, Sug_{\text{narr}})$
        \Comment{Revision}
        \State $(\mathcal{P},\mathcal{S}) \gets \mathcal{A}_{\text{idea}}(I_p, I_n, \mathcal{F}, \mathcal{D}_r)$
        \State $\mathcal{PR} \gets \mathcal{A}_{\text{gen}}(I_p,\mathcal{P}, \mathcal{S})$; $\mathcal{C} \gets \mathcal{A}_{\text{gen}}(\mathcal{PR}, I_p)$

        \State \textbf{continue}
    \EndIf

    \If{$Score_{\text{photo}} < \tau_{\text{photo}}$}
        \State $Sug_{\text{photo}} \gets \mathcal{A}_{\text{crit}}(\mathcal{C}, Score_{\text{photo}})$
        \Comment{Refinement}
        \State $(\mathcal{P},\mathcal{S}) \gets \mathcal{A}_{\text{idea}}(I_p, I_n, \mathcal{F}, \mathcal{D}_r, Sug_{\text{photo}})$
        \State $\mathcal{PR} \gets \mathcal{A}_{\text{gen}}(I_p,\mathcal{P}, \mathcal{S})$; $\mathcal{C} \gets \mathcal{A}_{\text{gen}}(\mathcal{PR}, I_p)$
    \Else
        \State \textbf{break} \Comment{Both gates satisfied}
    \EndIf
\EndFor

\State \Return $\mathcal{C}^* \gets \mathcal{C}$

\end{algorithmic}
\end{algorithm}

\section{Experiments}
\subsection{Implementing Details}



\paragraph{Models and Baselines}

We instantiate our pipeline using \texttt{GPT-4o} for multimodal understanding and reasoning, and evaluate four image generation models: \texttt{Nano Banana Pro}, \texttt{Nano Banana}, \texttt{Seedream~4.5}, and \texttt{Flux Kontext Max}. All generators are integrated without model-specific tuning. As baselines, we adopt direct generation where images are produced in a single pass conditioned only on the product packshot and name (with an optional reference grid), omitting narrative ideation, structured critique, and closed-loop refinement. All our qualitative results use \texttt{Nano Banana Pro} as the backbone.

\paragraph{Evaluation Metrics}
Evaluating grid collage imagery differs fundamentally from evaluating single images.
A grid collage should be evaluated as a unified visual composition, with explicit consideration of cross-grid relationships. Accordingly, we adopt a campaign-level, multi-criteria evaluation protocol.
We employ the MLLM \texttt{GPT-5.2} to score generated grid collages along three dimensions:
\textbf{Aesthetics}, measuring color harmony, compositional hierarchy, grid balance, and lighting quality;
\textbf{Richness}, capturing function coverage, information density, and product relevance across grids;
and \textbf{Coherence}, assessing product-centric narrative consistency, identity alignment, stylistic consistency, and scene coherence.
To complement the MLLM-based aesthetic assessment, we additionally report ArtiMuse~\cite{cao2025artimuse} scores as an external aesthetic metric, providing an independent measure of visual appeal.
Detailed scoring criteria are provided in Appendix~\ref{appx_visual quality scoring prompt}.

\begin{figure}
\centering
\includegraphics[width=1\linewidth]{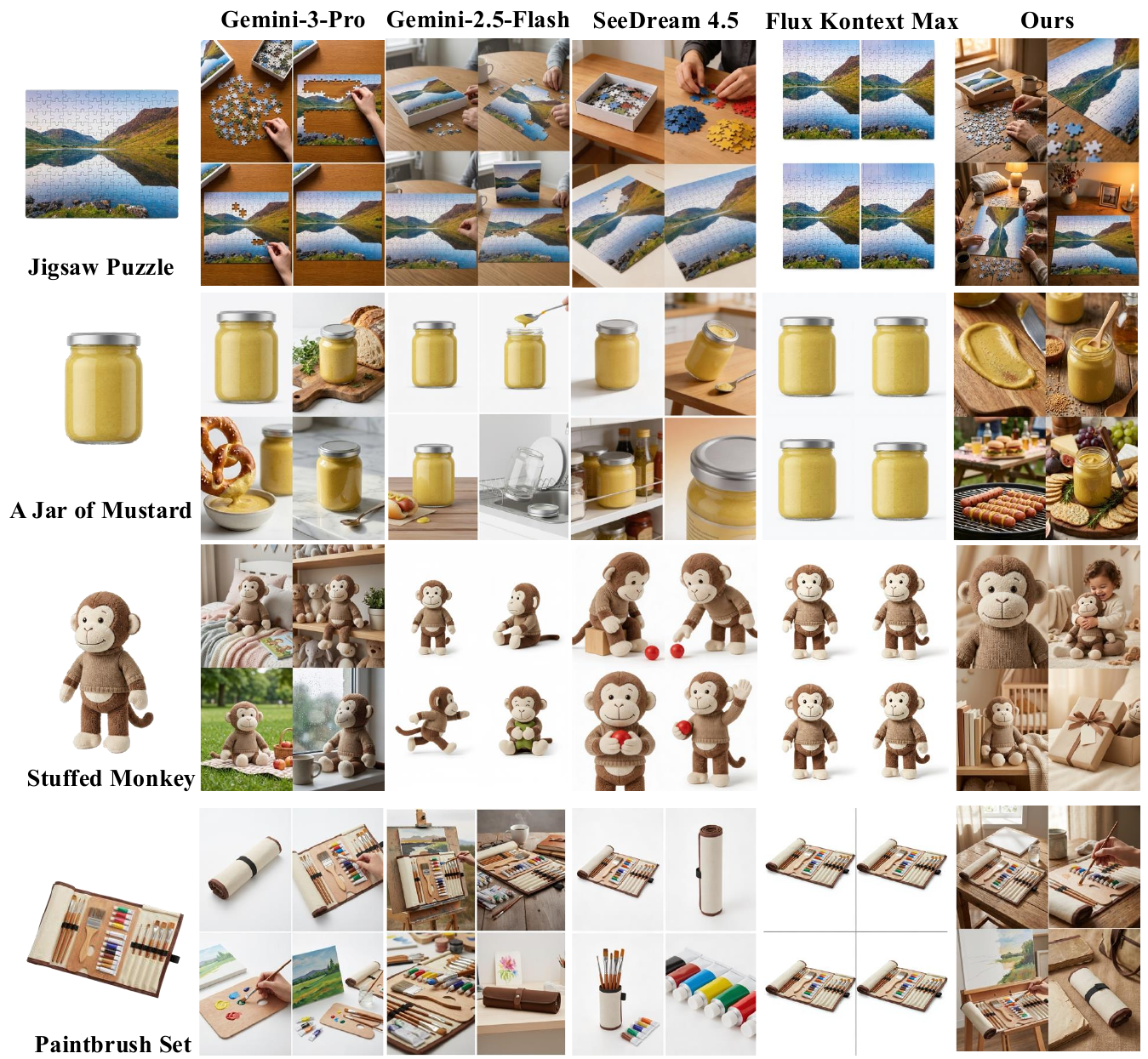}
\caption{Comparison between our method and  direct generation baselines under \textbf{creation mode}. Results are evaluated on 400 product images, with up to 3 iterations.}
\vspace{-3mm}
\label{fig_base_noref}
\end{figure}



\begin{table}[t]
\centering
\caption{MLLM-based evaluation in creation mode (averaged over four generation models and 400 test prompts).}
\label{tab_creation_vlm}
\setlength{\tabcolsep}{4pt}
\scalebox{0.7}{
\begin{tabular}{lcccc}
\toprule
Model &
\makecell{ArtiMuse$\uparrow$} &
\makecell{Aesthetic$\uparrow$} &
\makecell{Richness$\uparrow$} &
\makecell{Coherence$\uparrow$} \\
\midrule

Nano Banana Pro   & 62.640 & 7.186 & 7.156 & 8.003 \\
+Ours             & \textbf{63.464} & \textbf{7.512} & \textbf{7.237} & \textbf{8.077} \\
\addlinespace[3pt]

Nano Banana       & 62.732 & 7.066 & 6.750 & 7.822 \\
+Ours             & \textbf{65.115} & \textbf{7.485} & \textbf{7.179} & \textbf{8.033} \\
\addlinespace[3pt]

Seedream 4.5      & 62.367 & 6.860 & 6.650 & 7.532 \\
+Ours             & \textbf{64.666} & \textbf{7.234} & \textbf{7.006} & \textbf{7.747} \\
\addlinespace[3pt]

Flux Kontext Max  & 56.384 & 6.444 & 4.777 & 7.544 \\
+Ours             & \textbf{64.118} & \textbf{7.157} & \textbf{7.023} & \textbf{7.908} \\

\bottomrule
\end{tabular}
}
\end{table}

\subsection{Creation Mode Evaluation}

\paragraph{\textbf{Qualitative Evaluation}}
As shown in Figure~\ref{fig_base_noref}, direct generation baselines often produce grids that are visually plausible in isolation but lack coherent narrative structure as a whole, resulting in limited storytelling capacity. 
A frequent failure mode is shallow semantic understanding. As illustrated in the ${2}^{nd}$ row (A Jar of Mustard), baselines capture the product’s appearance but fail to explore its functional context, usage scenarios, or semantic variations across grids.
In other cases, grids appear visually plausible but lack meaningful semantic connections. As shown in the ${3}^{rd}$ row (Stuffed Monkey), the results of Nano Banana Pro manifest as independently generated images that are assembled into a collage without clear transitions or coherent progression.

In contrast, our agentic pipeline mitigates these issues through explicit \textit{product narrative framework}. By jointly reasoning about \textit{What to Shoot} and \textit{How to Shoot}, the system introduces deliberate viewpoint variation, strengthens inter-grid semantic connections, and enables usage-driven narrative associations. Thus, the generated grids evolve from isolated images into coherent campaign collages that consistently convey product identity and function.



\begin{figure}
\centering
\includegraphics[width=1\linewidth]{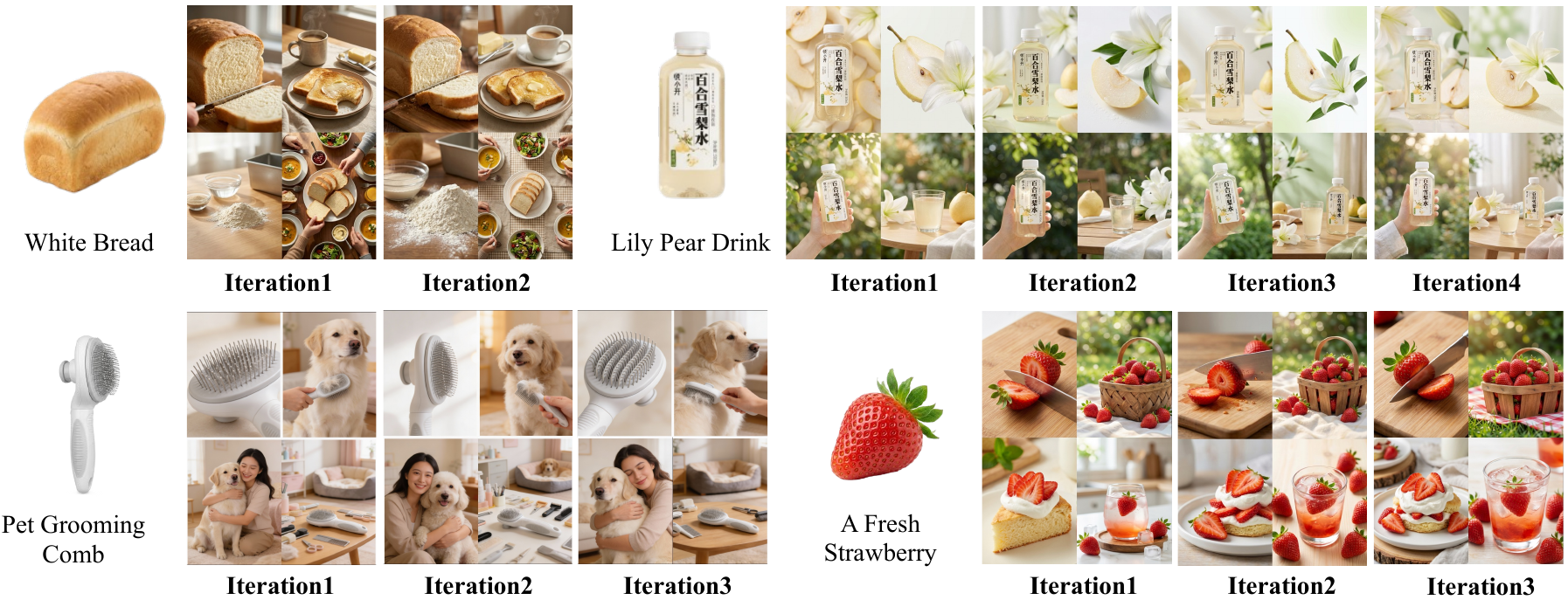}
\caption{Results of iterative refinement under \textbf{creation mode}.
Experiments are conducted with a maximum of five refinement iterations, and terminate early when the generated grid satisfies the predefined evaluation thresholds.}
\vspace{-5mm}
\label{fig_iteration_noref}
\end{figure}

\paragraph{\textbf{Quantitative Evaluation}}
Table~\ref{tab_creation_vlm} shows quantitative results comparing our agentic pipeline against direct generation baselines. We report ArtiMuse scores and MLLM-based evaluations on aesthetic quality, richness, and coherence.
The consistent improvements across different backbone generators indicate that the performance gains are attributable to the proposed agentic pipeline, rather than to model-specific tuning.

\begin{table*}[t]
\centering
\caption{Reference-mode evaluation: visual quality, reference transfer, and structural similarity.}
\label{tab_refmode_allinone_pluscka_last}
\setlength{\tabcolsep}{4pt}
\scalebox{0.85}{
\begin{tabular}{lcccccccc}
\toprule
Model &
\multicolumn{4}{c}{Visual Quality} &
\multicolumn{3}{c}{Reference Transfer} &
\multicolumn{1}{c}{Structural Similarity} \\
\cmidrule(lr){2-5}\cmidrule(lr){6-8}\cmidrule(lr){9-9}
&
\makecell{ArtiMuse\\Score$\uparrow$} &
\makecell{Aesthetic\\Score$\uparrow$} &
\makecell{Richness\\Score$\uparrow$} &
\makecell{Coherence\\Score$\uparrow$} &
\makecell{Photography\\Transfer$\uparrow$} &
\makecell{Narrative\\Transfer$\uparrow$} &
\makecell{Product\\Adaptation$\uparrow$} &
\makecell{CKA$\uparrow$} \\
\midrule

Nano Banana Pro
 & 58.375 & 7.123 & 5.559 & 7.588 & 7.239 & 6.981 & 6.665 & 0.907 \\
\quad +Ours
 & \textbf{61.160} & \textbf{7.529} & \textbf{6.103} & \textbf{7.813} & \textbf{7.462} & \textbf{7.171} & \textbf{8.203} & \textbf{0.934} \\
\addlinespace[1pt]

Nano Banana
 & 59.184 & 7.101 & 5.328 & 6.788 & 7.119 & 6.624 & 6.273 & 0.880 \\
\quad +Ours
 & \textbf{63.581} & \textbf{7.534} & \textbf{5.946} & \textbf{7.822} & \textbf{7.138} & \textbf{7.010} & \textbf{8.113} & \textbf{0.930} \\
\addlinespace[1pt]

Seedream 4.5
 & 59.164 & 6.866 & 5.177 & 6.744 & 6.817 & 6.325 & 6.367 & 0.872 \\
\quad +Ours
 & \textbf{64.499} & \textbf{7.321} & \textbf{5.886} & \textbf{7.584} & \textbf{7.371} & \textbf{7.255} & \textbf{8.362} & \textbf{0.932} \\
\addlinespace[1pt]

Flux Kontext Max
 & 59.114 & 6.494 & 4.550 & 5.708 & 3.376 & 3.252 & 5.185 & 0.714 \\
\quad +Ours
 & \textbf{62.933} & \textbf{7.240} & \textbf{5.744} & \textbf{7.802} & \textbf{6.696} & \textbf{6.548} & \textbf{7.958} & \textbf{0.921} \\

\bottomrule
\end{tabular}

}
\end{table*}

\paragraph{\textbf{Iteration Analysis}}

\begin{figure}
\centering
\includegraphics[width=0.8\linewidth]{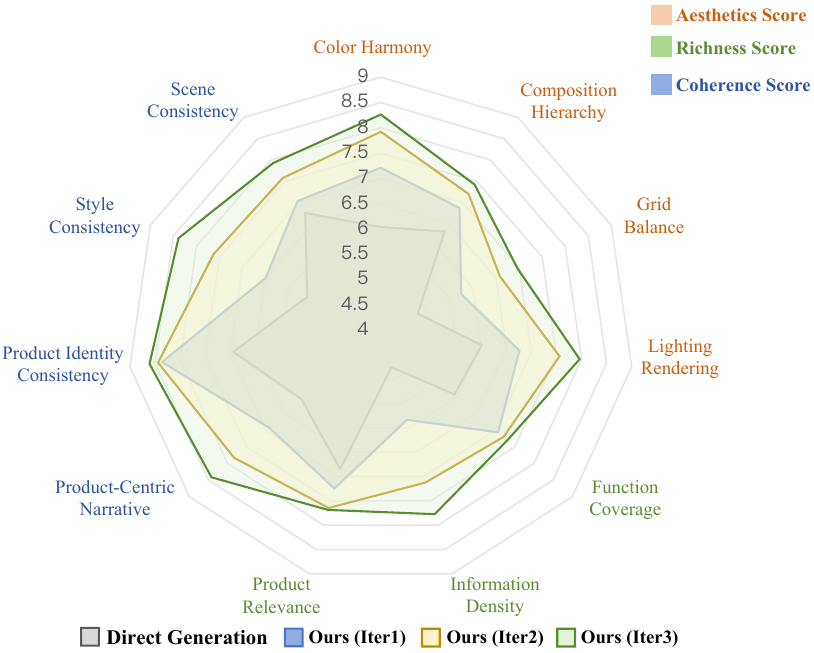}
\caption{  Iteration evaluation under \textbf{creation mode}.
We report scores of \textit{direct generation} and our iterative outputs after 1/2/3 refinement rounds (\textit{Iter1--Iter3}).}
\vspace{-4mm}
\label{fig_iter_quantity}
\end{figure}

Figure~\ref{fig_iteration_noref} illustrates the effect of increasing refinement iterations.
Figure~\ref{fig_iter_quantity} provides a fine-grained analysis of iterative refinement by visualizing MLLM scores over 3 dimensions (aesthetic, richness, coherence) and 11 sub-dimensions. 
Scores generally increase across iterations, with particularly clear gains on richness and coherence. This indicates that the proposed iterative pipeline is effective, providing a reliable mechanism to improve narrative richness and visual coherence.

\paragraph{Multi-Panel Grid Generation}

{
\setlength{\textfloatsep}{4pt plus 2pt minus 2pt}
\begin{figure}
\centering
\includegraphics[width=1\linewidth]{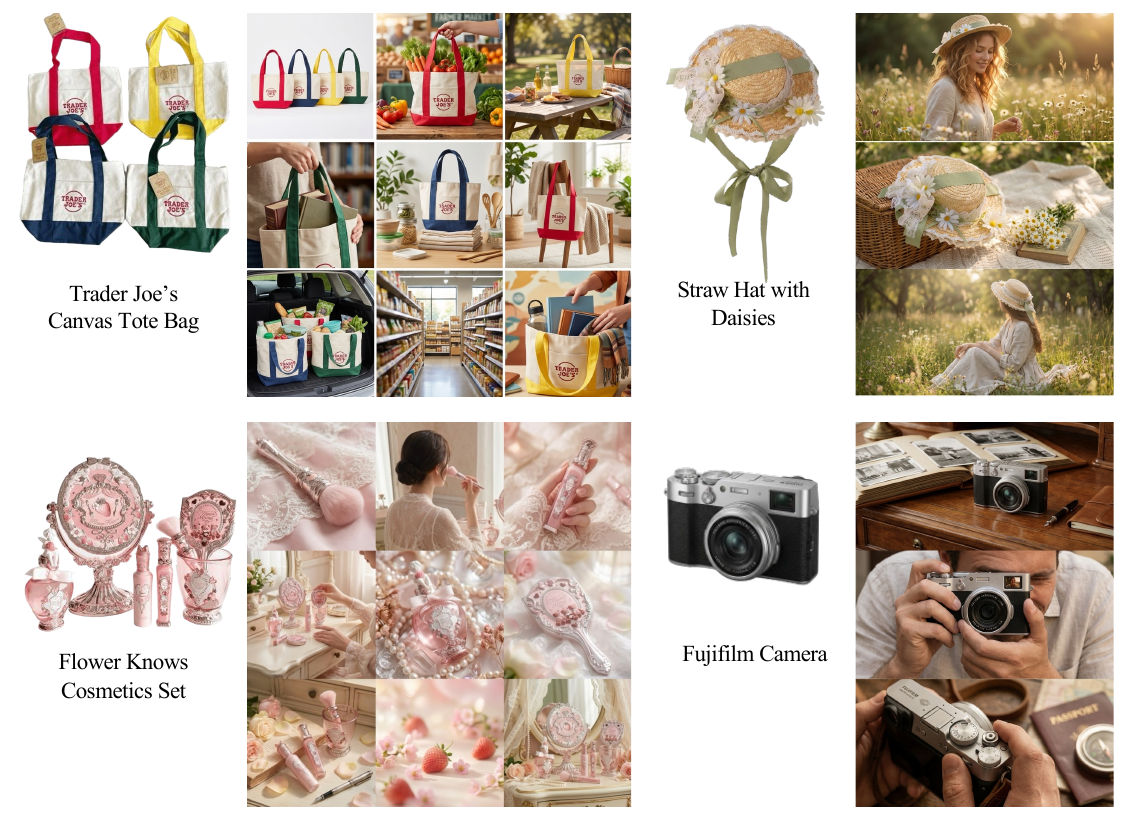}
\caption{Our framework supports diverse layouts (e.g., 3$\times$3 and 1$\times$3), enabling flexible grid collages.}
\vspace{-4mm}
\label{fig_multipanel}
\end{figure}
}

We extend our pipeline beyond 2$\times$2 grid layouts to flexible multi-panel compositions with varying grid counts and aspect ratios. As shown in Figure~\ref{fig_multipanel}, the 3$\times$3 nine-grid layout suits social media platforms where grid-based storytelling is prevalent, enabling explicit presentation of spatial and semantic relationships among multiple subjects. The 1$\times$3 horizontal layout produces cinematic sequences emphasizing mood, continuity, and emotional progression. These results demonstrate that our method is layout-agnostic, automatically adapting its generation strategies to diverse structures while preserving narrative coherence and visual consistency.

\subsection{Reference Mode Evaluation}

\paragraph{\textbf{Qualitative Evaluation}}

\begin{figure}
\centering
\includegraphics[width=1\linewidth]{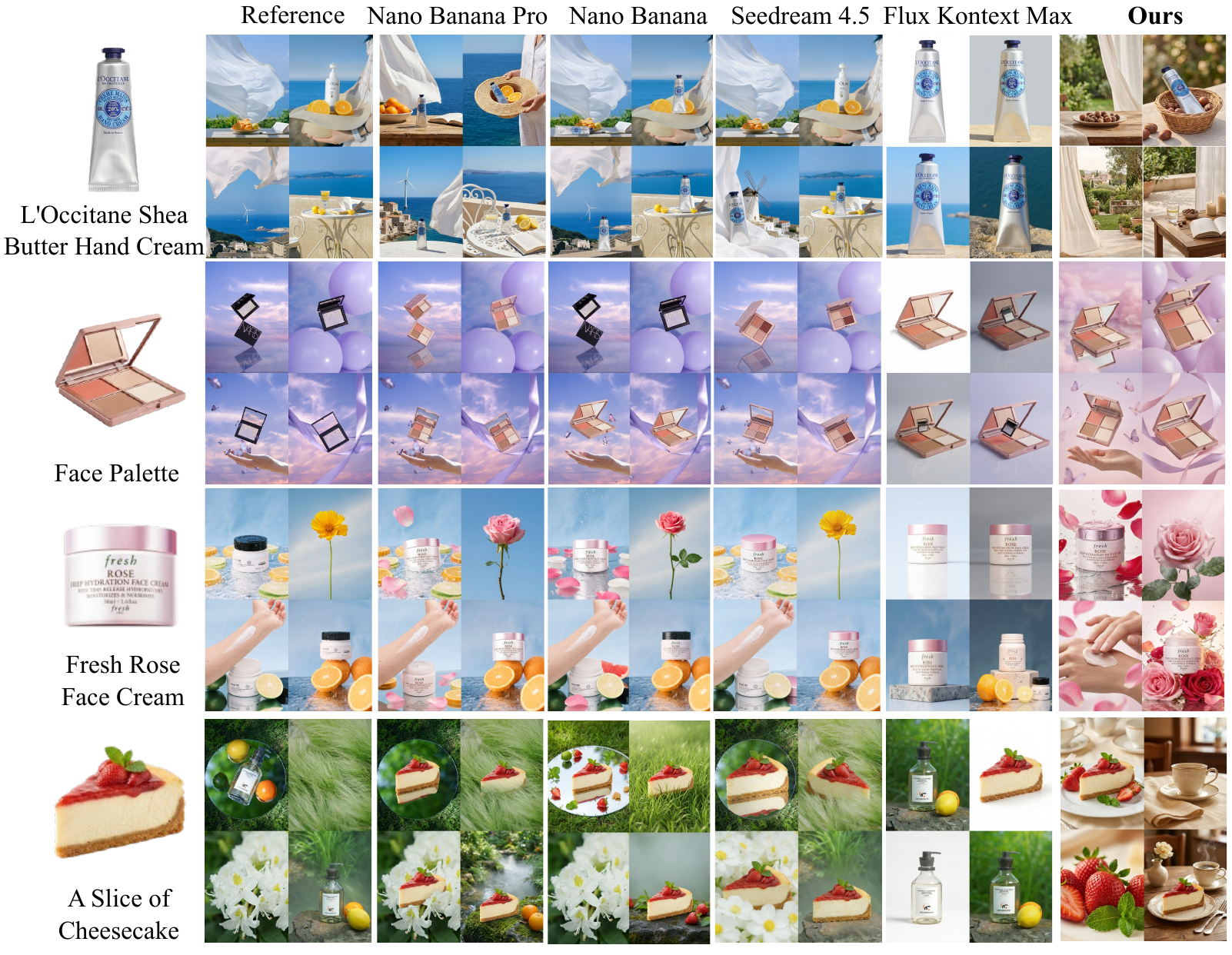}
\caption{Comparison between our method and direct generation baselines under \textbf{reference mode}. Results are evaluated on 500 reference-product pairs, with up to 3 iterations.}
\vspace{-4mm}
\label{fig_base_ref}
\end{figure}

We curate a dataset of 500 professionally photographed four-grid collages as visual references. Each reference collage is paired with a target product, forming 500 reference–product pairs.
Figure~\ref{fig_base_ref} qualitatively illustrates typical failure modes of direct baselines.
They frequently reuse reference semantics without reinterpretation, leading to attribute mismatches. For example, in the ${2}^{nd}$ row (Face Palette), the baseline inherits the reference color tone, inconsistent with the specific product. Similarly, sensory cues such as flavor or scent are often transferred incorrectly (e.g., ${1}^{st}$ row \textit{L’Occitane Shea Butter Hand Cream}), resulting in semantically implausible narratives. In more severe cases, incompatible context environments are transferred (e.g., ${4}^{th}$ row, A Slice of Cheesecake).
In contrast, our method captures the underlying campaign intents and reinterprets them according to the target product’s attributes.

\paragraph{\textbf{Quantitative Evaluation}}

In reference mode, the goal is not only to generate visually appealing collages but to faithfully adapt the narrative and compositional structure of a reference image to a target product.
Accordingly, we conduct a multi-level evaluation that assesses
(1) visual quality,
(2) structural alignment,
and (3) MLLM-based similarity.

\paragraph{Visual Quality}
We first evaluate the generated collages independently of the reference, using MLLM-based scores same as the creation mode.
Table~\ref{tab_refmode_allinone_pluscka_last} assesses whether reference guidance preserves strong advertising quality in terms of visual polish, informational coverage, and campaign-level consistency.

\paragraph{Structural Alignment}




Effective reference-guided generation should align with the reference at the structural level rather than through appearance copying. We evaluate reference adherence at the \emph{inter-grid structural level}. Each $2\times2$ campaign grid is represented by an \emph{inter-grid relation matrix} computed from pairwise CLIP similarities. 
We measure the alignment between the reference and generated relational structures using \emph{Centered Kernel Alignment (CKA)}~\cite{kornblith2019similarity}. Results are shown in Table~\ref{tab_refmode_allinone_pluscka_last}.



\paragraph{MLLM-based Similarity}

We further introduce an MLLM-based Reference Score,
which evaluates whether the \emph{semantic intent} of the reference is correctly transferred, whether the generated collage successfully adapts the reference’s grid planning and narrative logic to the target product, without copying reference-specific appearance.
Specifically, we report scores on \textit{Photography Transfer}, \textit{Narrative Transfer}, and \textit{Product Adaptation}, where higher values indicate better transfer. Results are shown in Table~\ref{tab_refmode_allinone_pluscka_last}.
Detailed scoring criteria are provided in Appendix~\ref{appx_reference transfer scoring prompt}.


\subsection{User Study}

We conduct an online user study ($N{=}83$) evaluating three aspects: \textbf{(i) realism} against professional advertisements, \textbf{(ii) visual appeal} in creation mode, and \textbf{(iii) reference-transfer} in reference mode. For realism, participants distinguish our results from real photography via pairwise forced choice. Our outputs are selected as ``real'' 48.3\% vs.\ 51.7\% for professional ads across 18 pairs, indicating participants cannot reliably distinguish them. For creation-mode appeal, participants choose which collage better motivates purchase; our method receives 71.8\% of votes (536/747) across 18 pairs. For reference-mode transfer, participants select which candidate better follows the reference composition while adapting to the target product; our method receives 76.1\% of votes (505/664) across 16 pairs, suggesting better reference learning without rigid copy-paste.
\begin{figure}[h]
    \centering
   \includegraphics[width=0.8\linewidth]{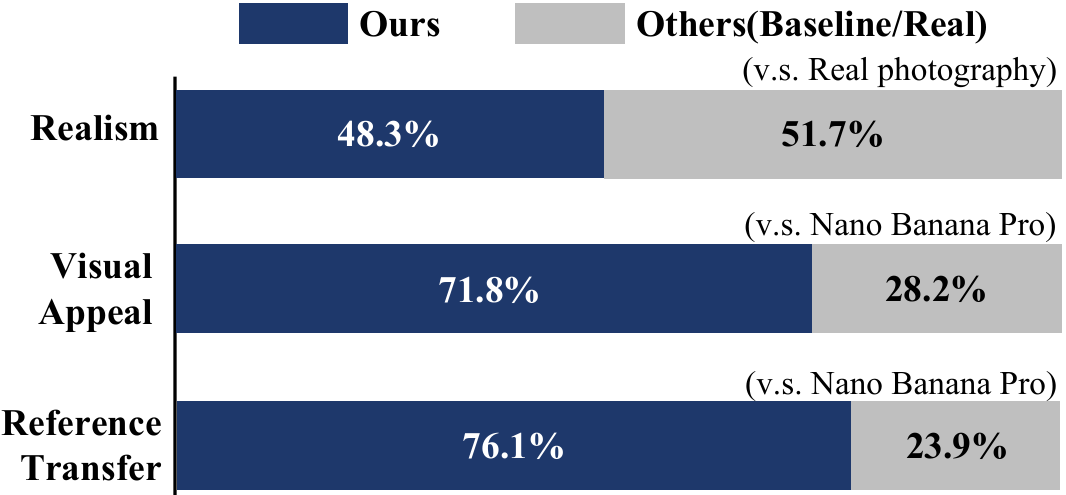}

    \caption{Results of the user study.}
    \label{fig_user}
    \vspace{-2mm}
\end{figure}



\subsection{Ablation and Limitations}

\begin{figure}[h]
    \centering
   \includegraphics[width=\linewidth]{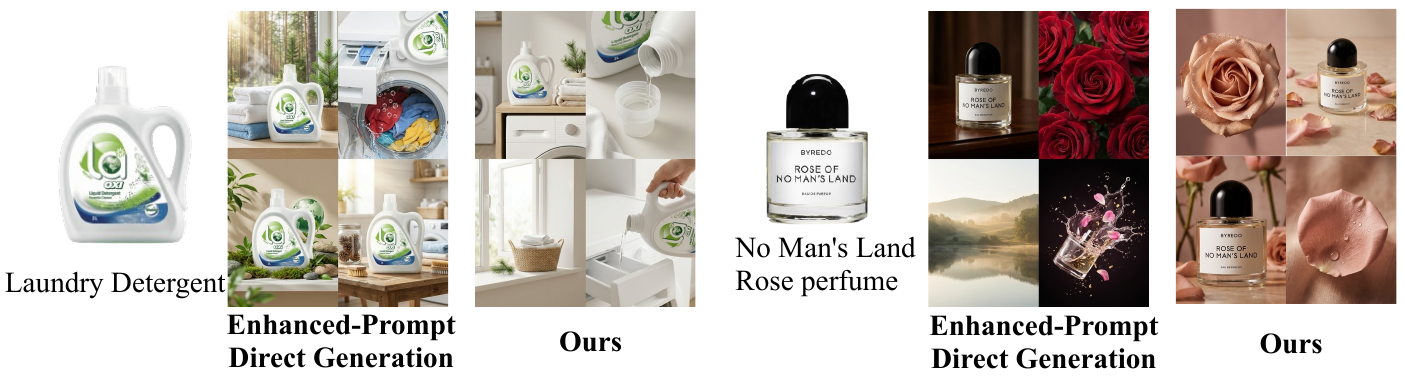}
    \caption{Comparison with enhanced-prompt direct generation.}
    \label{fig_ablation}
\end{figure}

\paragraph{Ablation Study}
To isolate the contribution of our \textit{Ideation-Generation-Critique} loop, we conduct an ablation that removes the $\mathcal{A}_{\text{idea}}$ and $\mathcal{A}_{\text{crit}}$.
Specifically, we use MLLM-models to obtain a brief understanding of the input product packshot, then write a set of prompts based on it, and finally generate the grid collage via the generation model (\emph{enhanced-prompt direct generation}).
As shown in Figure~\ref{fig_ablation}, this ablation verifies that the full \textit{Ideation-Generation-Critique} loop is effective: beyond prompt rewriting, it functions as a CoT-like reasoning process that improves overall visual quality.

\paragraph{Limitations}
As shown in Figure~\ref{fig_bad}, when generating product images for two goods from entirely distinct categories with minimal correlation, the results may exhibit visual incongruity. Additionally, our method occasionally generates unobserved product surfaces or internal structures, such as the prepared instant noodles depicted in Figure~\ref{fig_bad}(a), resulting in discrepancies between the generated imagery and the actual product appearance.
{
\setlength{\textfloatsep}{4pt plus 2pt minus 2pt}
\begin{figure}[h]
\centering
\includegraphics[width=1\linewidth]{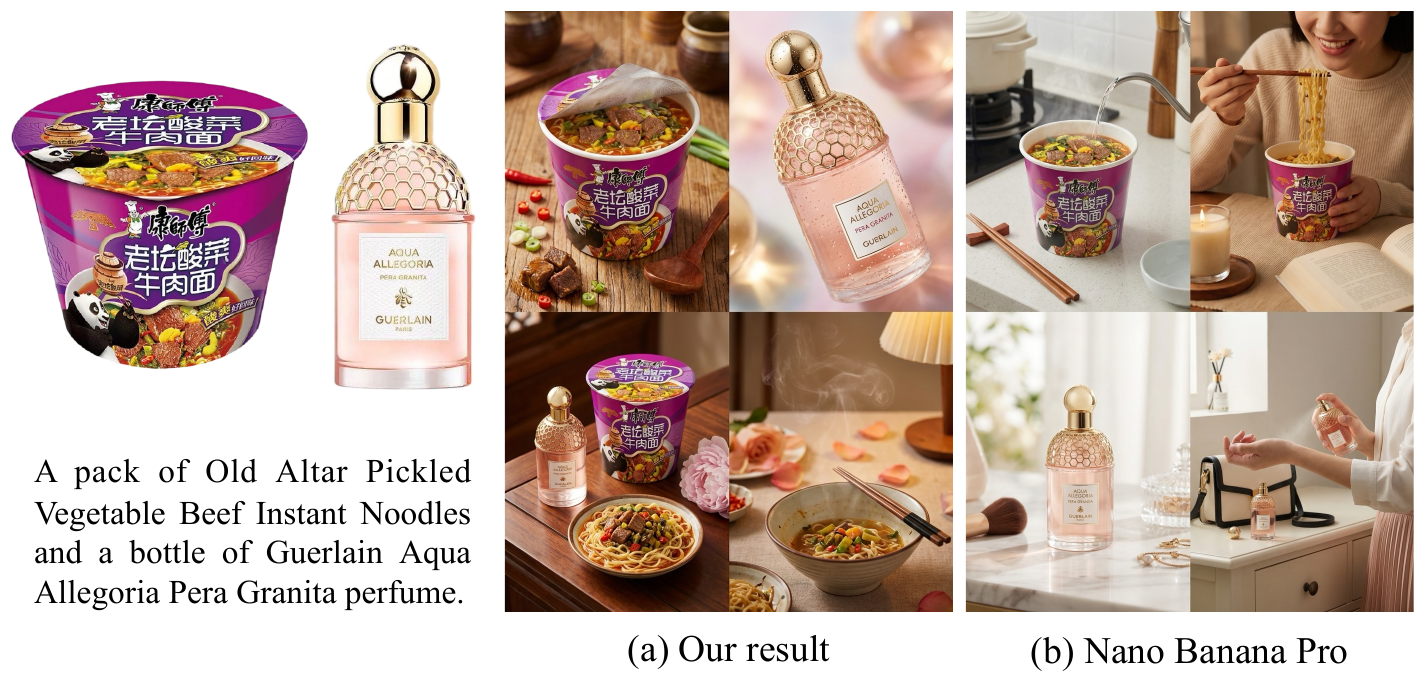}
\caption{Results for two products from distinct categories.}
\vspace{-4mm}
\label{fig_bad}
\end{figure}
}

\section{Conclusion}

In this work, we present a self-reasoning agentic framework that elevates product grid collage generation from simple depiction to coherent visual narrative. By establishing a structured \textit{Product Narrative Framework} and employing a gated, self-reflective critique loop, our method effectively bridges the gap between semantic understanding and pixel-level synthesis. 
Extensive evaluations demonstrate that our approach significantly outperforms direct generation baselines in narrative richness, visual coherence, and aesthetic quality across diverse grid layouts. Future work will explore extending this agentic paradigm to video storytelling and integrating 3D priors to further enhance structural fidelity.
\section*{Impact Statement}
This paper advances agentic generation of multi-grid collages, aiming to improve narrative richness, visual coherence, aesthetic quality and controllable reference-based transfer.

The approach may reduce the cost of campaign prototyping and enable faster creative iteration for designers and small businesses. However, the same capability can be misused to produce misleading advertisements or unauthorized brand-style imitations. We recommend responsible deployment with human review in commercial settings and clear disclosure/provenance when generated imagery is used. Future work should further address safeguards against deceptive claims and unauthorized reuse.

\bibliography{AgProImg-ICML}
\bibliographystyle{icml2026}

\newpage
\appendix
\onecolumn

\section{User Study Details}
\label{app:user_study}

\paragraph{Participants}
We collected $N{=}83$ valid responses via an online questionnaire platform.
Among participants, 56.6\% report advertising/design-related industry experience, 20.5\% are professionally trained, and 22.9\% have no relevant background.
Participants report moderate familiarity with AI-generated advertisements (mean 3.55/5).

\paragraphautorefname{Protocol}
All stimuli are $2{\times}2$ product campaign collages.
The study is within-subject.
Questions are presented in randomized order, and the left/right placement of candidates is randomized to mitigate position bias.
Participants answer all questions with mandatory responses.

\paragraph{Task 1: Realism (indistinguishability from real ads).}
Participants are shown \textbf{18 comparison pairs}, each consisting of one professionally shot advertisement grid and one grid generated by our method.
They answer a forced-choice question:
\emph{``Which one looks like the real (photographer-shot) advertisement?''}
We report vote splits per pair and also aggregate votes across all pairs.
Overall, our outputs are selected as ``real'' \textbf{361} times versus \textbf{386} times for professionally shot ads (48.3\% vs.\ 51.7\%), indicating that participants cannot reliably distinguish our results from real photography.
At the per-pair level, the vote splits are concentrated around chance: \textbf{most pairs fall within $\pm10\%$ of a 50/50 split}. The most indistinguishable pair is almost perfectly split (50.6\% vs.\ 49.4\%).

\paragraph{Task 2: Visual appeal in creation mode.}
To evaluate visual appeal and purchase motivation, participants complete forced-choice judgments on \textbf{18 comparison pairs} between our method and direct-generation baselines:
\emph{``Which grid better motivates you to buy this product?''}
We aggregate votes across all pairs. Our method receives \textbf{536} out of \textbf{747} votes (71.8\%),
demonstrating consistently stronger perceived attractiveness than direct generation.

\paragraph{Task 3: Reference-mode transfer ability.}
In reference mode, participants are provided with a reference grid and two candidates generated for the target product (ours vs.\ direct baseline),
and make forced-choice judgments on \textbf{16 comparison pairs}.
They answer:
\emph{``Which candidate better learns the reference’s composition/narrative while remaining adapted to the target product (i.e., not a rigid copy-paste)?''}
Across all pairs, our method receives \textbf{505} out of \textbf{664} votes (76.1\%).
This supports that our approach achieves \emph{flexible} transfer: capturing the reference’s layout and storytelling logic while preserving product-specific semantics.

\paragraph{Discussion and Limitations}
Overall, the user study supports our claims from three complementary perspectives:
(i) realism approaching indistinguishability from professional ads in the best cases,
(ii) improved aesthetics in creation mode,
and (iii) stronger reference-mode composition/narrative transfer.
Limitations include an online setting, self-reported participant background, and a finite number of evaluated product/reference pairs; nonetheless, the consistent preference trends align with our quantitative and qualitative results in the main paper.

\section{Structural Alignment}

\begin{figure}[h]
\centering
\includegraphics[width=0.55\linewidth]{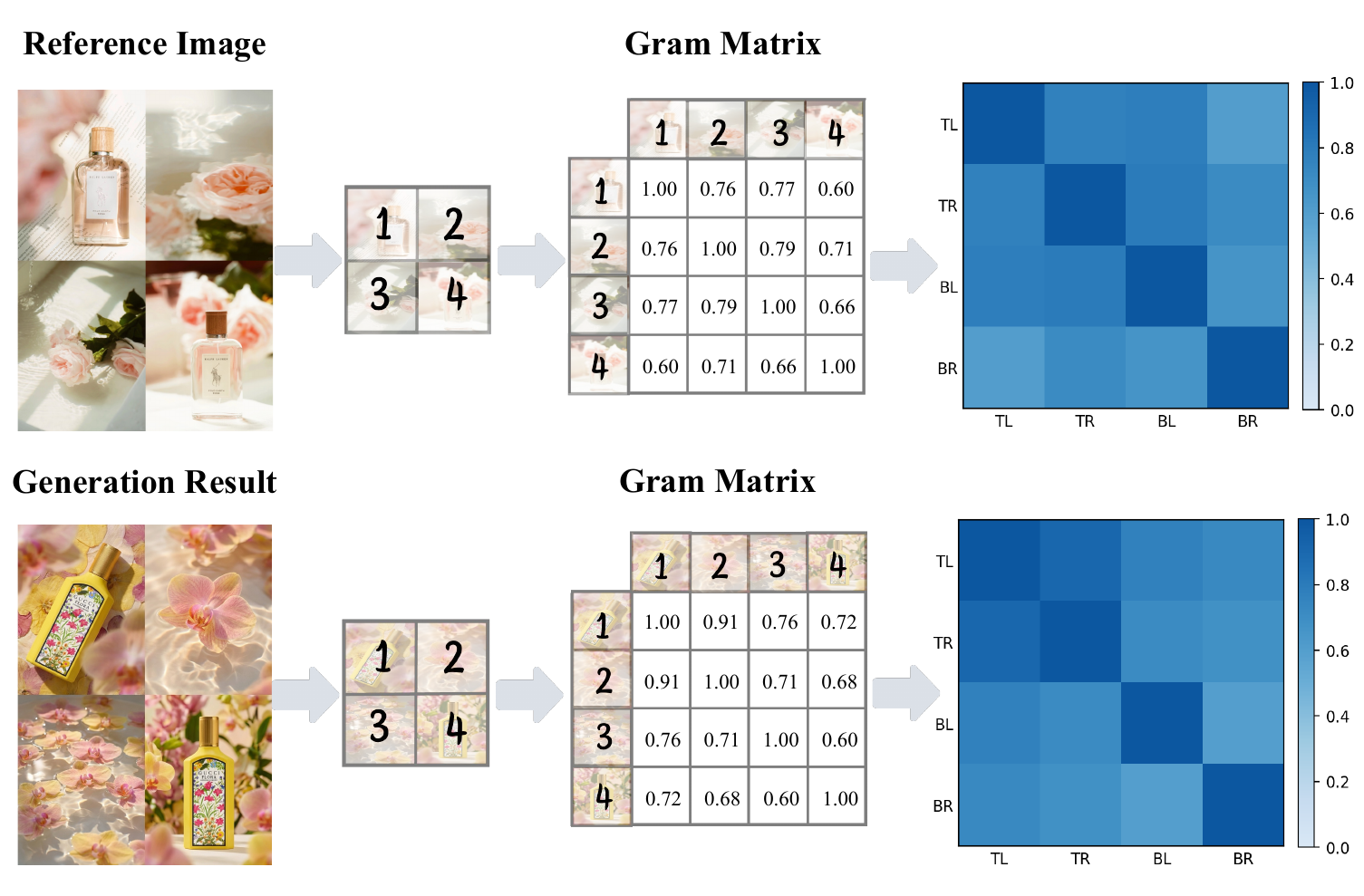}
\caption{Visualization of inter-grid relation matrices.}
\label{fig_reference_visualization}
\end{figure}

Effective reference-guided generation should follow the \emph{structural plan} of the reference campaign grid (i.e., how panels relate to each other)
rather than trivially copying appearance.
To quantify reference adherence at the \emph{inter-panel relational level}, we represent each $2{\times}2$ campaign grid as an
\emph{inter-grid relation matrix} derived from pairwise semantic similarities between its four panels.
Figure~\ref{fig_reference_visualization} shows representative relation matrices for a reference grid and a generated grid.

\subsection{Panel Embeddings and Relation Matrix}
Given a grid $G$ with four panels $\{x_1,x_2,x_3,x_4\}$ (ordered as TL, TR, BL, BR), we extract a CLIP embedding for each panel.
Concretely, we use the CLIP image encoder to obtain feature vectors
$\{f_i\in\mathbb{R}^{d}\}_{i=1}^{4}$, and apply $\ell_2$ normalization:
\begin{equation}
e_i = \frac{f_i}{\|f_i\|_2}, \qquad E = 
\begin{bmatrix}
e_1^\top \\ e_2^\top \\ e_3^\top \\ e_4^\top
\end{bmatrix}
\in \mathbb{R}^{4\times d}.
\end{equation}

We then construct the relation matrix as the Gram matrix of cosine similarities:
\begin{equation}
R = EE^\top \in \mathbb{R}^{4\times 4}, \qquad
R_{ij} = \langle e_i, e_j\rangle.
\end{equation}
Here, $R_{ij}$ captures the semantic proximity between panels $i$ and $j$ (e.g., shared scene context, product attributes, or narrative continuity),
and thus encodes higher-order campaign structure beyond per-panel realism or pixel-level similarity.

\subsection{Centered Kernel Alignment (CKA)}
To compare the relational structures of a reference grid $G^{\mathrm{ref}}$ and a generated grid $G^{\mathrm{gen}}$,
we compute their relation matrices $R^{\mathrm{ref}}$ and $R^{\mathrm{gen}}$ and evaluate their alignment using
\emph{Centered Kernel Alignment (CKA)}.
CKA measures similarity between two kernel (Gram) matrices, and is invariant to isotropic scaling of the matrices, which makes it robust to
global similarity magnitude differences across grids.

We first apply centering to each relation matrix:
\begin{equation}
\tilde{R} = HRH, \qquad
H = I - \tfrac{1}{4}\mathbf{1}\mathbf{1}^\top ,
\end{equation}
where $I$ is the $4\times 4$ identity matrix and $\mathbf{1}$ is the all-ones vector.
The CKA score is then:
\begin{equation}
\mathrm{CKA}(R^{\mathrm{ref}}, R^{\mathrm{gen}})
=
\frac{\langle \tilde{R}^{\mathrm{ref}}, \tilde{R}^{\mathrm{gen}} \rangle_F}
{\|\tilde{R}^{\mathrm{ref}}\|_F \, \|\tilde{R}^{\mathrm{gen}}\|_F},
\end{equation}
where $\langle A,B\rangle_F = \mathrm{tr}(A^\top B)$ denotes the Frobenius inner product and $\|\cdot\|_F$ is the Frobenius norm.
By construction, $\mathrm{CKA}(\cdot,\cdot)\in[-1,1]$; in practice, since $R$ is a cosine-similarity Gram matrix, scores are typically non-negative,
and higher values indicate stronger structural alignment.

\subsection{Interpretation}
Intuitively, high CKA implies that the \emph{pattern of pairwise relations} among panels in the generated grid matches that of the reference grid.
For example, if the reference has a strong semantic linkage between TL and TR (e.g., both are ``hero''-type shots) and a weaker linkage to BR
(e.g., a contextual or outcome panel), then a structurally aligned generation should reproduce a similar relational pattern, even if the exact imagery differs.
This makes CKA a suitable metric for assessing whether the model learns the reference grid's \emph{layout narrative and role structure}
instead of copying superficial textures.

\section{Creation Agent Architecture}
\label{app:creation_agent}

We implement \textbf{creation mode} as an iterative, three-agent pipeline that generates a $2{\times}2$ campaign collage from a single product packshot.
The overall objective is to produce a four-panel campaign that (i) preserves \textbf{packshot product identity}, (ii) forms a \textbf{coherent campaign world},
and (iii) expresses a \textbf{purchase-relevant narrative} through coordinated panel roles.
The pipeline alternates between \emph{planning} (text), \emph{image generation} (vision), and \emph{critique-driven refinement} (vision+text),
and terminates when the critique gates pass or when reaching a preset iteration budget.

\paragraph{Inputs and State.}
Each run takes as input (1) an authoritative white-background \textbf{product packshot} and (2) a short \textbf{user description} of marketing intent.
The system maintains an explicit state containing intermediate plans, compiled prompts, generated images, and critique results.
For reproducibility, we persist intermediate artifacts to disk, including JSON plans, compiled prompts, and generated collages.

\subsection{Agent 1: Ideation Orchestrator (Plan $\rightarrow$ Plan $\rightarrow$ Prompts)}
Agent~1 converts the packshot and user intent into a structured four-panel campaign plan, then compiles this plan into per-panel generation prompts.
It contains three staged modules:

\paragraph{Stage 1 (WHAT-level Planner).}
\textbf{Goal:} decide \emph{what to communicate} about the product, jointly modeling product attributes and target consumer logic.
The planner outputs a \texttt{product\_narrative\_framework} with five fields:
\emph{product\_essence} (what it is), \emph{product\_usage} (how it is used),
\emph{usage\_context} (when/where it appears), \emph{target\_consumer\_profile} (who and why),
and a coherent \emph{narrative\_framework} describing a shootable campaign story.
\textbf{Hard rules:} no surreal elements and no invented product capabilities; all elements must remain commercially shootable and purchase-relevant.

\paragraph{Stage 2 (HOW-level Four-Grid Visual Planner).}
\textbf{Goal:} translate the Stage~1 narrative into a \emph{four-panel photographic plan} plus a \emph{global visual style}.
For each panel (TL/TR/BL/BR), the planner specifies:
\emph{shot\_scale} (macro/close/medium/wide),
\emph{hero\_presence} (full/partial/none) and \emph{hero\_number},
\emph{subject\_emphasis} (primary/secondary subjects, attention direction, focus/DoF, separation strategy),
\emph{spatial\_composition}, and \emph{interaction} (product--environment logic).
Additionally, it outputs a \texttt{global\_visual\_style} describing \emph{color}, \emph{lighting}, \emph{style} (realism/production value),
and \emph{emotion\_mood}, which must remain consistent across all panels.

\paragraph{Stage 3 (Prompt Compiler).}
\textbf{Goal:} compile the structured Stage~2 plan into four concise natural-language prompts
\texttt{top\_left}, \texttt{top\_right}, \texttt{bottom\_left}, \texttt{bottom\_right}.
\textbf{Constraint:} the compiler is instructed to \emph{not introduce new ideas} and to faithfully translate the plan into generation-ready prompts,
while consistently reflecting the global visual style in every panel prompt.

\subsection{Agent 2: Image Generator (Packshot-Conditioned $2{\times}2$ Collage)}
Agent~2 takes the compiled four prompts and the packshot, and generates a \emph{single} $2{\times}2$ collage using an image model in an
\emph{image-edit / reference-conditioned} mode (the packshot is provided as the authoritative identity anchor).
The generator enforces two global instruction blocks:

\paragraph{Packshot Fidelity Constraint.}
The prompt explicitly states that the packshot is ground truth and requires:
(1) identical product identity (silhouette, materials, label structure, colorway),
(2) no added/removed parts or invented logos/text,
and (3) physically plausible rendering (no floating, clipping, impossible reflections).
If a scene description conflicts with fidelity, fidelity takes priority.

\paragraph{Campaign Aesthetic Constraint.}
The prompt further encourages a unified campaign world: consistent palette and lighting,
clear focal hierarchy per panel, minimal supportive props, clean backgrounds, and varied shot scales across the four panels.

The generated collage is saved to \texttt{output\_dir} with an iteration-indexed filename, and the path is stored in the shared state.

\subsection{Agent 3: Critique and Rerun Controller (Two-Gate Feedback)}
Agent~3 evaluates the generated collage against two sequential critique gates and decides whether to rerun ideation stages.

\paragraph{Gate 1 (WHAT-level Critique; 0--5 each).}
Gate~1 assesses conceptual correctness only (ignoring aesthetics): whether the collage communicates
\emph{product\_essence}, \emph{product\_usage}, \emph{usage\_context}, and \emph{target\_consumer\_profile}.
If any dimension falls below a threshold (default: $\ge 4$), the system triggers a rerun of \textbf{Stage~1} (and downstream stages),
passing back targeted revision suggestions.

\paragraph{Gate 2 (HOW-level Critique; 0--5 each).}
If Gate~1 passes, Gate~2 evaluates execution quality:
\emph{photographic realism}, \emph{campaign coherence}, and \emph{advertising aesthetic quality}.
If any dimension falls below a threshold (default: $\ge 4$), the system triggers a rerun of \textbf{Stage~2} and \textbf{Stage~3}
(with prompt-level, directly actionable edit suggestions), while keeping the Stage~1 narrative intact.

\paragraph{Iterative Refinement Loop.}
In each iteration, the pipeline executes:
\[
\text{(Ideation: Stage 1/2/3)} \;\rightarrow\; \text{(Generation)} \;\rightarrow\; \text{(Critique: Gate 1/2)}.
\]
A failed Gate~1 causes a \emph{concept-level} revision (rerun Stage~1--3), while a failed Gate~2 causes an \emph{execution-level} revision
(rerun Stage~2--3).
All critique outputs (scores, failed dimensions, and suggestions) are stored as structured JSON to support ablations and reproducibility.

\section{Reference Agent (Reference Mode)}
\label{app:reference_agent}

In \textbf{reference mode}, we introduce a \emph{Reference Agent} that conditions generation on a given $2{\times}2$ reference campaign grid.
Rather than copying appearance, the agent extracts \textbf{transferable decisions} from the reference—i.e., high-level, product-agnostic
layout and narrative choices—and uses them to guide the target product’s collage generation.

\paragraph{Input and Objective.}
The agent receives (i) a \textbf{reference grid} $G^{\mathrm{ref}}$ and (ii) the \textbf{target product packshot} (plus optional user intent).
Its objective is to distill a set of structured, reusable design decisions that preserve the reference’s \emph{campaign plan} while remaining
faithful to the target product identity.

\paragraph{Transferable Decisions.}
The extracted decisions are organized into two levels:
(1) \textbf{What to shoot (narrative framework)}: an abstract, product-agnostic campaign story distilled from the reference grid,
including the core message progression across panels (e.g., \emph{context} $\rightarrow$ \emph{product essence} $\rightarrow$ \emph{usage/action}
$\rightarrow$ \emph{benefit/mood wrap-up}), the intended consumer appeal logic, and the semantic role of each quadrant (TL/TR/BL/BR) without inheriting
reference-specific objects or brands;
and (2) \textbf{How to shoot (photography decisions + global visual style)}: per-panel photographic directives such as shot scale
(macro/close/medium/wide), viewpoint and focal hierarchy, subject--background separation, prop/environment interaction patterns, and a shared
global style block specifying lighting setup, color palette, texture/contrast treatment, background cleanliness, and overall emotional tone.
These decisions serve as transferable constraints to guide generation while allowing the concrete visual content to be re-instantiated for the target product.

\paragraph{Anti-copy Constraints.}
To prevent shallow imitation, the agent is instructed to treat reference elements (props, scenes, metaphors) as \emph{abstract templates} and to
re-instantiate them in a product-appropriate way for the target category (e.g., preserving ``macro texture close-up'' as a decision, but changing
the concrete surface/ingredient to match the target product).
Literal reproduction of reference objects, brand marks, or category-specific props that conflict with the target product is penalized in the critique stage.

\paragraph{Output and Usage.}
The agent outputs a structured \texttt{transfer\_plan} that specifies per-panel decisions (TL/TR/BL/BR) and a global style block.
This plan is then fed into the same prompt compiler and generator as creation mode, enabling reference-guided generation that aligns in
\emph{structure and narrative logic} without appearance copying.
\section{MLLM-based Visual Quality Scoring Prompt}
\label{appx_visual quality scoring prompt}

We employ a Multimodal Large Language Model (MLLM) as an automatic evaluator for each generated $2{\times}2$ campaign collage.
The VLM takes four panels (top-left, top-right, bottom-left, bottom-right) as input and outputs structured scores along three axes:
\textbf{Aesthetics}, \textbf{Richness}, and \textbf{Coherence}. Each axis is decomposed into several sub-dimensions (below).
All scores are integers in a fixed range (1--10), accompanied by brief evidence-based rationales.

\paragraph{Prompt Summary (with Definitions and Scoring Guidelines)}
The evaluator is told that the input consists of four images forming a $2{\times}2$ product campaign grid (TL/TR/BL/BR).
It must score the grid on three axes—\textbf{Aesthetics}, \textbf{Richness}, and \textbf{Coherence}—and for each sub-dimension output
(i) an integer score in $[1,10]$ and (ii) a short evidence-based rationale. The prompt enforces \textbf{JSON-only} output.

\paragraph{Aesthetics (visual execution quality).}
\textbf{Definition:} Aesthetics measures visual beauty, appeal, and professional craft of the four panels, focusing strictly on \emph{form}
(composition, lighting/material rendering, color, and layout balance) rather than informational completeness.
\textbf{Constraint:} Aesthetics must \emph{not} consider product information coverage, narrative progression, or how much the viewer learns.
Minimal or simple images may still score highly if execution is strong.
\textbf{Sub-dimensions:} (A1) \emph{composition\_hierarchy} (framing, focal clarity, cropping, negative space, depth),
(A2) \emph{lighting\_rendering} (light direction coherence, shadows/highlights, tonal range, material realism),
(A3) \emph{color\_harmony} (palette sophistication, contrast/saturation control, subject--background relationship),
(A4) \emph{grid\_balance} (distribution of visual weight across the 2$\times$2 layout; balance across rows/columns/diagonals).
\textbf{Scoring guideline:} $9$--$10$ indicates professional, intentional art direction with clean framing and convincing lighting/materials;
$7$--$8$ indicates generally strong execution with minor issues (e.g., slightly generic composition or mild grading imbalance);
$5$--$6$ indicates acceptable but unrefined craft (e.g., mild clutter, flat light, or generic palette);
$3$--$4$ indicates clear visual problems (weak hierarchy, inconsistent light, unpleasant color, or lopsided layout);
$1$--$2$ indicates severe failure in visual logic (chaotic composition, broken lighting, or extreme imbalance).

\paragraph{Richness (product information coverage).}
\textbf{Definition:} Richness measures information density and complementary product coverage across the four panels, i.e.,
whether each panel adds \emph{distinct and meaningful} product-related content rather than repeating the same shot type.
\textbf{Constraint:} Richness evaluates informational breadth and depth \emph{not} narrative smoothness; a coherent story can still be low-richness if content is redundant.
\textbf{Sub-dimensions:} (R1) \emph{function\_coverage} (coverage of multiple campaign roles such as hero/feature/usage/detail/result/mood),
(R2) \emph{information\_density} (amount of substantive product-related information per panel; avoid sparse ``vibes-only'' panels),
(R3) \emph{product\_relevance} (whether props/scenes/metaphors are clearly tied to product attributes/benefits rather than generic filler).
\textbf{Scoring guideline:} $9$--$10$ indicates broad role coverage (typically $\ge$4 distinct functions) with high per-panel informativeness and minimal filler;
$7$--$8$ indicates good variety (3--4 functions) with small gaps or mild redundancy;
$5$--$6$ indicates moderate variety (2--3 functions) with noticeable repetition or sparse panels;
$3$--$4$ indicates mostly one repeated function and/or heavy reliance on generic atmosphere;
$1$--$2$ indicates almost no meaningful product information (near-total redundancy or unrelated filler).

\paragraph{Coherence (campaign unity and consistency).}
\textbf{Definition:} Coherence measures whether the four panels form a unified campaign with consistent product identity, product-serving intent,
consistent visual world/style, and a non-contradictory campaign logic.
\textbf{Sub-dimensions:} (C1) \emph{product\_identity\_consistency} (same product/variant cues; stable shape/color/branding),
(C2) \emph{product\_centric\_narrative} (each panel must clearly serve the product via presence, usage, result, ingredient, or product-tied metaphor),
(C3) \emph{style\_tone\_consistency} (no medium mixing; consistent grading, texture, realism level, and mood),
(C4) \emph{world\_campaign\_cohesion} (shared world/campaign intent with complementary, non-contradictory progression).
\textbf{Special rule (hero-less panels):} panels without visible product are allowed \emph{only if} they are unambiguously product-serving
(e.g., showing results/outcomes, ingredients/components, sensory cues, usage context, or a strongly product-tied metaphor).
\textbf{Scoring guideline:} $9$--$10$ indicates clear same-product identity, consistent style pipeline, and a cohesive campaign world with complementary panels;
$7$--$8$ indicates mostly consistent identity/style with minor mismatches or small narrative gaps;
$5$--$6$ indicates partial cohesion with noticeable identity drift, style mismatch, or fragmented campaign logic but still somewhat related;
$3$--$4$ indicates strong fragmentation (unjustified product absence, medium/style clashes, or contradictory worlds);
$1$--$2$ indicates panels appear unrelated or inconsistent enough to be different products/campaigns.

\paragraph{Penalty Rules and Output Constraint.}
The evaluator is explicitly instructed to \textbf{heavily penalize} (i) unjustified panels that do not serve the product,
(ii) noticeable product identity drift without explanation, (iii) mixed visual medium (photo vs.\ 3D vs.\ illustration), and
(iv) abrupt world/campaign breaks without a unifying rationale. The final response must be \textbf{valid JSON only} that matches the predefined schema.

\paragraph{Output Schema}
\begin{lstlisting}[style=prompt]

"aesthetics": {"subdim": {"score": 1-10, "reason": "..."} , ...},
"richness":   {"subdim": {"score": 1-10, "reason": "..."} , ...},
"coherence":  {"subdim": {"score": 1-10, "reason": "..."} , ...}

\end{lstlisting}

\section{MLLM-based Reference Transfer Scoring Prompt}
\label{appx_reference transfer scoring prompt}
In addition to absolute visual quality scoring, we also use an MLLM to evaluate \emph{reference-conditioned transfer quality} between a
\textbf{REFERENCE\_GRID} and a \textbf{GENERATED\_GRID}, both provided as full $2{\times}2$ grid images.
The evaluator is explicitly instructed to \textbf{not crop} either image, and to assess structural transfer with \textbf{position-wise alignment}
(\texttt{TL(ref)}$\leftrightarrow$\texttt{TL(gen)}, \texttt{TR(ref)}$\leftrightarrow$\texttt{TR(gen)}, \texttt{BL(ref)}$\leftrightarrow$\texttt{BL(gen)}, \texttt{BR(ref)}$\leftrightarrow$\texttt{BR(gen)}),
while judging narrative/story holistically across all four panels.
The prompt emphasizes \emph{structural and logical transfer} rather than literal object copying, and penalizes shallow imitation.

\paragraph{Axis A: Grid Plan Transfer (Hard Constraints).}
\textbf{Definition:} Measures whether the generated grid reproduces the reference grid’s \emph{panel-level plan} at corresponding positions,
including (i) \emph{hero presence and role} (product-present vs.\ product-absent; full/partial/no hero; number of products),
(ii) \emph{shot scale} (macro/close/medium/wide), (iii) \emph{subject emphasis} (primary vs.\ secondary elements; focus/DoF cues),
(iv) \emph{spatial composition} (placement, relative size, negative space, visual weight),
and (v) \emph{interaction pattern} (product--prop/environment relation; or, for product-absent panels, whether context implies the product).
\textbf{Special rule:} if the reference grid contains any product-absent panel, the generated grid is expected to preserve an equivalent
product-absent panel serving a similar role; filling all panels with product is penalized.
\textbf{Scoring guideline (1--10):} $9$--$10$ strong positional alignment across most panels with preserved rhythm;
$7$--$8$ minor deviations but overall plan learned; $5$--$6$ partial alignment with clear drift in some positions;
$3$--$4$ weak alignment; $1$--$2$ no meaningful plan transfer.

\paragraph{Axis B: Narrative Logic Transfer (Soft Constraints).}
\textbf{Definition:} Measures whether the generated grid learns the reference grid’s \emph{abstract campaign story logic} rather than copying literal content.
The evaluator considers: (i) \emph{story-world abstraction} (product essence/sensory identity, usage actions, lifestyle context),
(ii) \emph{narrative progression} (e.g., context$\rightarrow$cues$\rightarrow$transformation$\rightarrow$hero; raw$\rightarrow$refined; problem$\rightarrow$solution),
(iii) \emph{campaign coherence} (one unified campaign vs.\ four isolated images), and
(iv) \emph{consumer attraction logic} (desire/curiosity/emotional resonance comparable to the reference).
\textbf{Scoring guideline (1--10):} $9$--$10$ clear, coherent logic transfer; $7$--$8$ mostly consistent with small gaps;
$5$--$6$ vague/incomplete resemblance; $3$--$4$ fragmented story; $1$--$2$ no recognizable shared logic.

\paragraph{Axis C: Product Fit to the Generated Grid (Anti-copy / Adaptation).}
\textbf{Definition:} Measures whether the generated grid remains faithful to \emph{its own product identity} and is plausibly adapted,
rather than performing a naive ``product swap'' that preserves the reference background/props/story even when implausible.
The evaluator checks category plausibility of materials and usage context, and whether scenes feel intentionally designed for the generated product.
\textbf{Scoring guideline (1--10):} $9$--$10$ product-specific and well-integrated; $7$--$8$ mostly appropriate with minor generic elements;
$5$--$6$ acceptable but weakly grounded; $3$--$4$ awkward/mismatched integration; $1$--$2$ clear reference-copy failure (swap without adaptation).

\paragraph{Output Constraint and Diagnostics.}
The evaluator must return \textbf{JSON only} with (i) the three transfer scores above (each with a short rationale),
(ii) \textbf{per-position} alignment notes for TL/TR/BL/BR (e.g., \texttt{strong/partial/weak}),
(iii) a short list of \texttt{key\_matches} and \texttt{key\_mismatches}, and (iv) an overall \texttt{verdict} (\texttt{pass/borderline/fail}).

\end{document}